\documentclass[journal,twoside,web]{ieeecolor}
\usepackage{tmi}
\usepackage{cite}
\usepackage{amsmath,amssymb,amsfonts}
\usepackage{algorithmic}
\usepackage{graphicx}
\usepackage{textcomp}
\usepackage{pifont}
\usepackage{multirow}
\usepackage{bm}
\usepackage{makecell}
\def\BibTeX{{\rm B\kern-.05em{\sc i\kern-.025em b}\kern-.08em
    T\kern-.1667em\lower.7ex\hbox{E}\kern-.125emX}}
\begin{document}
\title{UTSRMorph: A Unified Transformer and Superresolution Network for Unsupervised Medical Image Registration}
\author{Runshi Zhang, Hao Mo, Junchen Wang, \IEEEmembership{Member, IEEE}, Bimeng Jie, Yang He, Nenghao Jin, Liang Zhu
	\thanks{This work was supported by the National Key Research and Development Program of China (Grant No. 2022YFC2405401), the Natural Science Foundation of China (Grant No. 62173014 and Grant No. U22A2051), the Natural Science foundation of Beijing Municipality (Grant No. L232037). (Corresponding author: Junchen Wang)}
	\thanks{Junchen Wang, Runshi Zhang, Hao Mo are with the School of Mechanical Engineering and Automation, Beihang University, Beijing, China (e-mail: zhangrunshi@buaa.edu.cn, wangjunchen@buaa.edu.cn).}
	\thanks{Bimeng Jie and Yang He are with Peking University School and Hospital of Stomatology, Beijing, China.}
	\thanks{Nenghao Jin and Liang Zhu are with Department of Stomatology, The First Center of Chinese PLA General Hospital, Beijing, China.}
}
\maketitle

\begin{abstract}
Complicated image registration is a key issue in medical image analysis, and deep learning-based methods have achieved better results than traditional methods.
The methods include ConvNet-based and Transformer-based methods.
Although ConvNets can effectively utilize local information to reduce redundancy via small neighborhood convolution, the limited receptive field results in the inability to capture global dependencies.
Transformers can establish long-distance dependencies via a self-attention mechanism; however, the intense calculation of the relationships among all tokens leads to high redundancy.
We propose a novel unsupervised image registration method named the unified Transformer and superresolution (UTSRMorph) network, which can enhance feature representation learning in the encoder and generate detailed displacement fields in the decoder to overcome these problems.
We first propose a fusion attention block to integrate the advantages of ConvNets and Transformers, which inserts a ConvNet-based channel attention module into a multihead self-attention module.
The overlapping attention block, a novel cross-attention method, uses overlapping windows to obtain abundant correlations with match information of a pair of images.
Then, the blocks are flexibly stacked into a new powerful encoder.
The decoder generation process of a high-resolution deformation displacement field from low-resolution features is considered as a superresolution process.
Specifically, the superresolution module was employed to replace interpolation upsampling, which can overcome feature degradation.
UTSRMorph was compared to state-of-the-art registration methods in the 3D brain MR (OASIS, IXI) and MR-CT datasets (abdomen, craniomaxillofacial).
The qualitative and quantitative results indicate that UTSRMorph achieves relatively better performance.
The code and datasets are publicly available at https://github.com/Runshi-Zhang/UTSRMorph.
%Editor: Please note that some text appears to be missing here. Please add any missing information.

\end{abstract}


\section{Introduction}
\label{sec:introduction}
\IEEEPARstart{D}{eformable} medical image registration~\cite{dmi} is a fundamental medical image analysis task.
It can estimate the dense nonlinear spatial correspondence between a moving image and a fixed image, and then transforms them into a common coordinate system~\cite{irr}.
The information that is fused from mono-/multimodality images can assist surgeons in diagnosing and performing operations, such as tumor invasion~\cite{tumor}, preoperative planning~\cite{preplan}, and intraoperative navigation~\cite{intra}.
Traditional methods~\cite{TomVercauteren,HeinrichMattiasP}  based on iterative computations view the optimization match problem with each pair of unseen images as a variational problem that calculates a smooth point-to-point mapping between a moving image and a fixed image. 
In traditional methods, the mapping model’s pattern is given as a priori (e.g., affine, spline, etc.), while in deep learning-based methods the mapping is represented by learnable deep network’s parameters, which makes the deep learning-based methods have a good representation learning capability. 

With rapid development, deep learning methods have provided many effective technical programs for medical image analysis tasks.
In particular, convolutional neural networks (ConvNets) and Transformer networks have performed well in the medical image community, such as intelligent diagnosis~\cite{diagnosis}, lesion identification~\cite{detection}, and segmentation~\cite{segmentation}.
Deep learning-based image registration methods~\cite{voxelmorphcvpr,voxelmorphtmi} have achieved better accuracy than traditional methods; additionally, they are several orders of magnitude faster.
The methods generate deformation fields by learning the feature representation of image registration pairs during training and the deformation results can be obtained quickly during testing.
The initial medical image registration methods rely on supervised ground-truth deformation fields, which are usually annotated by traditional registration methods.
Thus, the accuracy of traditional methods limits the performance of supervised methods, causing deformation errors and inferior generalizability.
G. Balakrishnan~\cite{voxelmorphcvpr} proposed an unsupervised learning framework based on ConvNets named VoxelMorph to generate a deformation field to overcome this problem.
Specifically, a pair of images is input to the VoxelMorph's UNet~\cite{RonnebergerOlaf}.
The feature representation is extracted by the UNet's encoder; the decoder can combine the multiscale features using skip-connection operations and interpolation upsampling layers to generate a high-resolution (HR) deformation field.
Then, a spatial transformation network~\cite{stn} can warp the moving image to a fixed image by relying on the deformation field.
Researchers have proposed several improved methods based on ConvNets~\cite{rdp,voxelmorphtmi,Learn2Reg}.
Although ConvNets can effectively extract local information and reduce redundancy within a small neighborhood, the limited receptive field is inherent.
ConvNets make it difficult for image registration methods to establish long-range information dependency, which represents the relations between one voxel and other far voxels.

Transformer networks~\cite{2017attention} are another essential deep learning method, and their self-attention mechanisms provide large receptive fields to capture long-range spatial information.
It establishes long-distance dependencies between each token by calculating the similarity of $query$-$key$; thus, Transformer could be suitable for registration tasks~\cite{transmatch}.
Vision Transformer (ViT)~\cite{vit} initially utilized a Transformer in computer vision (CV), achieving outstanding performance and extensive application.
The input image is split into fixed-size patches as tokens, and a learned linear embedding layer is used to encode them.
After adding a position embedding, a standard Transformer block is employed to calculate the relationship of each patch.
The intense calculation of the similarity of all tokens leads to high redundancy~\cite{likunchang}.
Swin Transformer~\cite{swint} (Swin-T) has significantly improved the computational efficiency and feature representation of Transformers by using shifted windows in CV. It also provides unbound cross-window designs.
In medical image registration, TransMorph~\cite{transmorph} based on Swin-T was proposed to obtain the displacement field, achieving better performance than VoxelMorph.
It benefits from the feature representation of Swin-T, and TransMorph still suffers from high redundancy.
Therefore, combining the advantages of ConvNets and ViTs is crucial to improving the representation learning ability of image registration methods.

In this paper, we propose a lightweight unified Transformer (ConvNet and Transformer) and superresolution (SR) method for medical image registration named UTSRMorph.
UTSRMorph uses and reconstructs encoder-decoder designs.
The encoder comprises our proposed fusion attention blocks (FAB) and overlapping attention blocks (OAB).
The encoder adopts the advantages of ConvNets and Transformers to balance long-distance information dependence and local information utilization.
It fully extracts the point-to-point relationships of image pairs through a cross-attention mechanism, enhancing its feature extraction ability.
In the decoder, the SR module is employed to handle the feature degradation of the encoder and generate a smoother and more detailed deformation field.
The main contributions of this paper are as follows:

(1) FAB, which combines the self-attention of ViTs and the channel attention of ConvNets, is proposed to enhance the feature representation learning ability.
It can accelerate optimization while reducing the local redundancy of our network based on Transformer.

(2) OAB, which is different from the nonoverlapping Swin-T windows, are designed using a novel cross-attention.
The abundant correlations with match information of image pairs are better integrated through cross-window connections. 

(3) The SR module is used to fuse multiscale features and generate detailed HR features.
It provides much richer and more meaningful feature representations from low-resolution (LR) features to the displacement field and reduces the computational cost.

To our knowledge, the lightweight modules which can fully integrate the advantages of ConvNets and ViTs, and adopting the SR approach in deformable medical image registration have not been previously reported.
Our proposed method is compared to state-of-the-art (SOTA) methods in two 3D brain MR datasets (OASIS, IXI), abdomen MR-CT and our craniomaxillofacial (CMF) tumor MR-CT datasets.
Our method achieves outstanding performance on several evaluation metrics and demonstrates its effectiveness.
\section{Related Work}
\subsection{Deep Learning-based Image Registration}
Deep learning-based image registration methods have been extensively studied in previous works and consist of ConvNet- and Transformer-based methods.
In recent years, ConvNet-based methods, which have achieved comparable accuracy to traditional methods, have significantly reduced the time cost.
G. Balakrishnan~\cite{voxelmorphcvpr} presented an unsupervised learning framework (VoxelMorph) for deformable medical image registration.
This approach achieved significant breakthroughs in terms of registration accuracy and computational costs on many datasets.
In their subsequent work~\cite{voxelmorphtmi}, organ segmentation information was used to improve VoxelMorph during training and significantly increased the registration accuracy.
B. Kim~\cite{cycle} proposed a cycle-consistent method to retain the original topology. This method can preserve the topology during deformation via implicit regularization.
H. Wang~\cite{rdp} relied on the recursive deformable pyramid (RDP) network, a pure convolutional pyramid, to handle large deformations.

Transformer-based methods using the self-attention mechanism can establish an interdependent relationship between each image pair window.
J. Chen~\cite{vitvnet} adopted simple bridging of ViT and V-Net (ViT-V-Net) to increase the registration accuracy of VoxelMorph.
In their later work~\cite{transmorph}, the Swin-T block was used as an encoder named TransMorph.
Z. Chen~\cite{transmatch} proposed TransMatch to input a pair of images separately into the encoder branch and extract features independently.
Additionally, the $query$ and $key/value$ windows were separately generated, and a cross-attention mechanism was used to calculate the cross-window correlations.
J. Zheng~\cite{RAN} proposed a novel motion separable backbone and residual aligner module to address complicated motion patterns by capturing separate motions and disentangling the predicted deformation across multiple neighboring organs. 

In summary, the above SOTA methods prove that the registration accuracy can be improved by a modified encoder, which is constructed by ConvNet- or Transformer-based layers.
Complicated upsampling fusion operations in the decoder are also commonly employed in these methods; however, complex designs can lead to GPU memory explosion.
Inspired by these methods, the integration of local features and long-distance dependence by ConvNets and Transformers is proposed in this paper. Additionally, a lightweight SR module is first used to handle complicated deformations.
The modules can generate an HR displacement field and decrease model complexity.
Our proposed method can increase feature representation learning ability and limit GPU memory usage when inputting large images.

\subsection{Combination of ConvNets and ViTs}
ConvNets and ViTs have been used to improve representation learning of vision backbones in recent works, such as position embedding based on ConvNets~\cite{CSWin} and replacing linear projection with convolutional projection in self-attention~\cite{wuhaiping}.
In the CV community, K. Yuan~\cite{yuankun} used convolutional patches stemming from input images to generate tokens, and a depthwise convolution layer was employed to insert a feedforward network into a multihead self-attention (MSA) module.
K. Li~\cite{likunchang} proposed UniFormer, which adopted dynamic position embedding and a multihead relation aggregator based on ConvNets to handle redundancy and dependency.
Similarly, X. Chen~\cite{HAT} proposed a method to combine ConvNet-based and self-attention modules.
In deformation medical image registration, ViT-V-Net~\cite{vitvnet} bridges ViT and ConvNets, yet simple bridging in the encoder layers does not effectively utilize the advantages of both.
In this paper, we combine ConvNets and ViTs to reduce the training cost significantly and achieve effective representation learning ability.
\subsection{Cross-Attention Mechanism}
The $query, key, value$ of the MSA module in Swin-T and ViTs are calculated separately from equal nonoverlapping windows, while the cross-attention mechanism adopts a more flexible window design.
Typically, $query$ and $key$-$value$ originate from different feature maps and then generate additional contextual information~\cite{crossattention1,crossattention2,crossattention3,crossattention4}.
Z. Peng~\cite{crossattention4} proposed an augmented cross-attention unit in a Conformer, which calculates the cross-attention of the ConvNet and Transformer branch features and enhances the classification and localization accuracy.
Y. Zheng~\cite{crossattention3} computed the cross-attention value between image patch features with context and structure information of the spatial relationship, achieving excellent performance in histopathology whole slide images (WSI) analysis.
In deformation medical image registration, TransMatch~\cite{transmatch} used cross-windows from moving images and fixed images to calculate cross-attention values and then realized explicit feature matching.
In this paper, a different cross-attention mechanism was proposed and inserted into the encoder.
Overlapping windows were used for each window feature with a large receptive field to calculate abundant correlations with matching information.
\subsection{Superresolution Upsampling}
The HR deformation displacement field is generated from LR features during the decoder, which is considered an SR process.
The downsampling layer of the backbone inevitably leads to feature degradation when transforming HR features to LR features.
This will be emphasized when restoring HR fine-grained features using an interpolation upsampling layer, such as bicubic interpolation.
Although interpolation-based upsampling was applied in early SR because of its simplicity, it usually generates much smoother results~\cite{moser}.
This may be detrimental to generating the deformation displacement field details.
Currently, learning-based upsampling is more popular.
Transposed convolution (TransConv) leads to crosshatch artifacts and fixed, redundant features because of zero padding~\cite{moser}.
To overcome this problem, ESPCN~\cite{pixel} was proposed to upscale the LR feature into the HR feature using a subpixel convolution layer.
Specifically, the layer contains a channel convolutional layer using an array of upscaling filters to obtain feature maps and a pixel-shuffle layer that can rearrange them to an HR output.
This approach has been applied in many medical image SR tasks\cite{pixeljbhi1,pixeljbhi2,pixeltmi}.
In this paper, we propose an SR module to generate fine-grained features and deformation fields for medical image registration.
\section{Method}
\begin{figure*}[!t]
\centerline{\includegraphics[width=\textwidth]{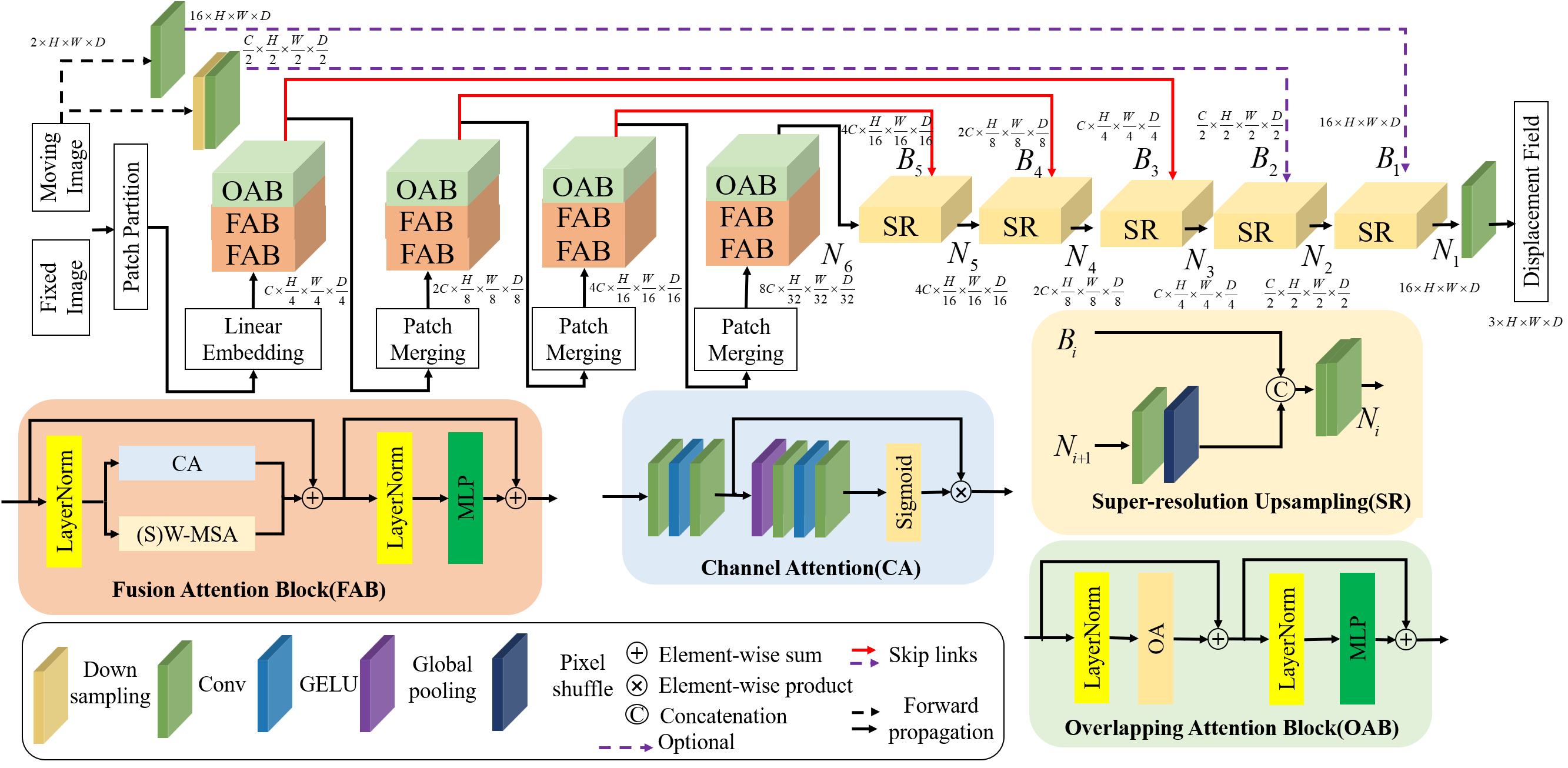}}
\caption{The unified Transformer and superresolution (UTSRMorph) network include four stages in the encoder and several SR modules in the decoder.
One stage consists of FABs and an OAB.
The black dashed arrows represent that the moving image and fixed image are concatenated as input to the following two optional convolutional blocks.}
	\label{fig1}
\end{figure*}
The framework for unsupervised medical image registration is described as~\cite{voxelmorphcvpr,voxelmorphtmi,transmatch}, and the optimal displacement field is written as
\begin{equation}\widehat{\phi}=\mathop{\arg\min}\limits_{\phi}\mathcal{L}_{sim}(f,m \circ \phi)+\lambda\mathcal{L}_{smooth}(\phi),\label{eq1}\end{equation}
where $f$ and $m$ are the fixed and moving images, respectively.
$\phi$ is the displacement field that maps $m$ to $f$.
$m \circ \phi$ represents the warped moving image using the displacement field.
The loss function $\mathcal{L}_{sim}(\cdot,\cdot)$ can measure the image similarity between the warped moving and fixed images.
$\mathcal{L}_{smooth}(\cdot)$, with a weight of $\lambda$, imposes regularization.

The key module in the registration framework is the network for obtaining the displacement field, which is the proposed unified Transformer and superresolution (UTSRMorph) network.
In this section, UTSRMorph is introduced in detail, and the overall pipeline is shown in Fig.\ref{fig1}.
UTSRMorph also adopts an architecture design that has been widely used in previous works~\cite{transmorph}; the structure is based on UNet~\cite{RonnebergerOlaf} and Swin-T~\cite{swint}.
The size of the moving and fixed images is $H\times W\times D$, and the input is $2 \times H\times W\times D$.
Based on ViT~\cite{vit} and Swin-T, the patch partition splits a 3D volume input into nonoverlapping patches treated as a "token".
The resolution and feature dimension of each patch are $P\times P\times P$ and $2 \times P\times P\times P$.
Then, a linear embedding layer is used to map to a feature map with arbitrary dimensions ($C$).
\begin{equation}\mathbf{z}_0=[x_p^1\mathbf{E};x_p^2\mathbf{E};...;x_p^N\mathbf{E}],\label{eq2}\end{equation}
where $x_p \in \mathbb{R}^{N\times(2P^3)}$ is the patch, and the lower case $p$ represents “patch”. 
$P=4$ in this paper.
$N=HWD/P^3$ is the resulting number of patches.
$\mathbf{E} \in \mathbb{R}^{2P^3\times C}$ is a learnable linear embedding layer, and the output is $\mathbf{z}_0 \in \mathbb{R}^{N\times C}$.

Then, four stages with several Transformer blocks are used on these patch tokens, and the blocks of each stage consist of several FABs and an OAB.
The linear embedding layer, FAB and OAB are referred to as "stage 1".
In other stages (the $i-th$ stage, $i=2, 3, 4$), a patch merging layer is added to concatenate the feature representation with $2\times 2\times 2$ neighboring patches. Then the number of tokens is reduced ($2\times$ downsampling of resolution) by $2\times 2\times 2=8$ to obtain $8D-$dimensional features $2^{(i+1)}C$.
The features are further input to a linear layer, and the output dimension is set to $2^{(i-1)}C$.
FAB uses local and global tokens to combine the advantages of ConvNets for decreasing local redundancy and ViTs for capturing long-range dependency via self-attention.
OAB uses an overlapping window instead of the Swin-T window to establish cross-window connections, enhancing the representative learning ability.
The features with unchanging resolution $H/2^{(i+1)}\times W/2^{(i+1)}\times D/2^{(i+1)}$ are calculated in these blocks.
Finally, four feature maps with the resolutions of $C \times \frac{H}{4}\times \frac{W}{4}\times \frac{D}{4}$, $2C \times \frac{H}{8}\times \frac{W}{8}\times \frac{D}{8}$, $4C \times \frac{H}{16}\times \frac{W}{16}\times \frac{D}{16}$, and $8C \times \frac{H}{32}\times \frac{W}{32}\times \frac{D}{32}$ are obtained.
Two optional convolutional blocks are used to generate two feature maps with HRs ($\frac{C}{2} \times \frac{H}{2}\times \frac{W}{2}\times \frac{D}{2}$, $16 \times H\times W\times D$) and then fully utilize the features with various resolutions.
Although convolutional blocks can slightly improve image registration accuracy according to~\cite{transmorph}, they consume considerable GPU memory.
The SR module is proposed in this paper to upsample features and uses a skip connection to fuse the HR features of the encoder.
The displacement field is obtained through the above equation.
The spatial transformation function~\cite{stn,transmorph} uses the displacement field to warp the moving image and generate a registration result.

In the following subsection, we first elaborate on FAB and OAB in the encoder.
Next, we detail how to upsample in the SR module of the decoder.
Finally, the loss functions in this network are introduced.
\subsection{Fusion Attention Block}
The standard Swin-T architecture uses window-based self-attention to establish relationships between one window (token) and all other windows, and the shifted window provides cross-window connections of nonoverlapping windows.
The MSA module treats different channel-wise features equally and calculates the relationship among the tokens, which can lead to high redundancy~\cite{likunchang}.
To overcome this problem, several previous works~\cite{wuhaiping,yuankun,likunchang} introduced the desirable properties of CNNs (shift, scale and distortion invariance) to ViTs.
Therefore, we propose FAB to combine MSA and convolutional channel attention (CA) into a standard Swin-T block.
The CA module transforms the channel-wise global spatial information into channel statistics, which are utilized to rescale the feature map, hence suppressing redundant information~\cite{channle}. 
By combining CA with MSA into FAB, our method can capture channel-wise dependencies effectively while aggregating local and global features, thus reducing redundancy. 
Specifically, the MSA and CA are inserted between two LayerNorm (LN) layers of the Swin-T block.
FAB uses the window-based multihead self-attention (W-MSA) module and shifted window-based self-attention (SW-MSA) module alternately, similar to Swin-T, and the shift size is set to half of the window size.
The standard CA module~\cite{channle} allows the module to focus on more informative features of all tokens and can extract the channel statistical information among channels.
The two FABs are formulated as
\begin{equation}\widehat{\mathbf{z}}^l=\text{W-MSA}(\text{LN}(\mathbf{z}^{l-1}))+\alpha \text{CA}(\text{LN}(\mathbf{z}^{l-1}))+\mathbf{z}^{l-1},\label{eq3}\end{equation}
\begin{equation}\mathbf{z}^l=\text{MLP}(\text{LN}(\widehat{\mathbf{z}}^l))+\widehat{\mathbf{z}}^l,\label{eq4}\end{equation}
\begin{equation}\widehat{\mathbf{z}}^{l+1}=\text{SW-MSA}(\text{LN}(\mathbf{z}^{l}))+\alpha \text{CA}(\text{LN}(\mathbf{z}^{l}))+\mathbf{z}^{l},\label{eq5}\end{equation}
\begin{equation}\mathbf{z}^{l+1}=\text{MLP}(\text{LN}(\widehat{\mathbf{z}}^{l+1}))+\widehat{\mathbf{z}}^{l+1},\label{eq6}\end{equation}  
where $\widehat{\mathbf{z}}^l$ and $\mathbf{z}^{l}$ are the features that are output by the (S)W-MSA and multilayer perceptron (MLP) modules for block $l$ of each stage, respectively.
$\alpha$ is a small multiplied weight of the CA module that regulates the optimization and visual representation of FAB.
The self-attention is formulated as
\begin{equation}\text{Attention}(Q,K,V)=\text{SoftMax}(QK^\text{T}/\sqrt{d}+B)V,\label{eq7}\end{equation}
where $Q,K,V\in \mathbb{R}^{P_xP_yP_z\times d}$ are computed by linear mappings named $query,key,value$.
$d$ is the $query/key$ dimension.
$B\in \mathbb{R}^{P_xP_yP_z\times P_xP_yP_z}$ is the relative position bias~\cite{transformer}.
$P_x,P_y,P_z$ is the patch size, and $P_xP_yP_z$ represents the number of patches in a 3D feature.
The range of the relative position is $[-\mathbf{P}+1,\mathbf{P}-1], \mathbf{P} = [P_x,P_y,P_z]^\text{T}$; then, a bias matrix $\widehat{B}\in \mathbb{R}^{(2P_x-1)\times (2P_y-1)\times (2P_z-1)}$ is parameterized, and the values in $B$ are taken from $\widehat{B}$.
Since the (S)W-MSA module also depends on the features of many token embedding channels to calculate self-attention and enhance visual representation, the features are directly computed in the CA module based on ConvNets, resulting in a high computational cost and information redundancy.
To solve this problem, we use two convolution layers to compress and recover the feature channel in the CA module with a compression parameter $\beta$.
\subsection{Overlapping Attention Block}
\begin{figure}[!t]
	\centerline{\includegraphics[width=\columnwidth]{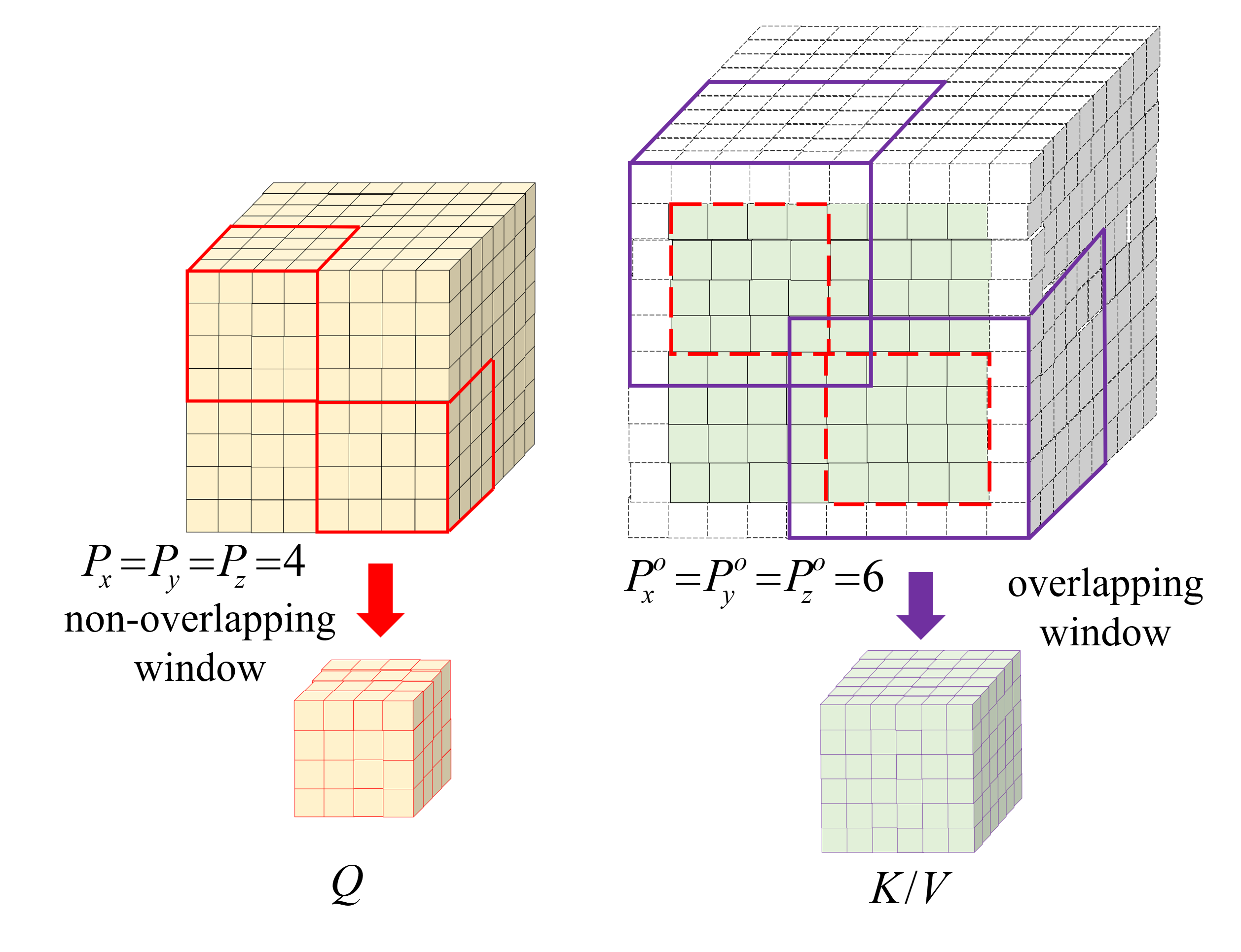}}
\caption{Overlapping window partition of OAB.}
	\label{fig2}
\end{figure}
The cross-attention mechanism~\cite{crossattention1,crossattention2,crossattention3,crossattention4,transmatch} is always employed in the context of Transformer architecture and allows more flexible multihead attention computations.
It can better integrate abundant correlations with visual information through cross-window connections.
Our proposed OAB is based on the standard Swin-T and is described as follows:
\begin{equation}\widehat{\mathbf{z}}^o=\text{OA}(\text{LN}(\mathbf{z}^{l+1}))+\mathbf{z}^{l+1},\label{eq8}\end{equation}
\begin{equation}\mathbf{z}^{l+2}=\text{MLP}(\text{LN}(\widehat{\mathbf{z}}^o))+\widehat{\mathbf{z}}^o,\label{eq9}\end{equation}
where $\widehat{\mathbf{z}}^o$ is the output of the OA module.
Specifically, $Q$ and $K/V$ come from nonoverlapping and overlapping windows, respectively, as shown in Fig.\ref{fig2}.
$I_Q,I_K,I_V$ are calculated by a linear map layer.
$I_Q$ is partitioned using $\frac{HWD}{P_xP_yP_z}$ local windows of size $P_x\times P_y \times P_z$, such as FAB, and then $Q$ is generated, while $\frac{HWD}{P_xP_yP_z}$ overlapping windows of size $P^{o}_x\times P^{o}_y \times P^{o}_z$ are used to unfold $I_K,I_V$, obtaining $K,V$.
The overlapping window size is defined as
\begin{equation}\mathbf{P}^o=(1+\epsilon)\times \mathbf{P},\label{eq10}\end{equation}
where $\epsilon$ is an overlapping parameter that can control the degree of window overlap.
$\mathbf{P} = [P_x,P_y,P_z]^\text{T}$ and $\mathbf{P}^o = [P^{o}_x,P^{o}_y,P^{o}_z]^\text{T}$ are the window sizes of $query$ and $key/value$, respectively.
This overlapping operation can be specifically described as the kernel size of the sliding partition with $P^{o}_x\times P^{o}_y \times P^{o}_z$, and the sliding stride with $P_x\times P_y \times P_z$ is used to partition features.
The features are padded to zero to ensure consistent partition windows.
The self-attention computing similarity is expressed as
\begin{equation}\text{Attention}(Q,K,V)=\text{SoftMax}(QK^\text{T}/\sqrt{d}+B^{o})V,\label{eq11}\end{equation}
The relative position bias $B^{o} \in \mathbb{R}^{P_xP_yP_z\times P^{o}_xP^{o}_yP^{o}_z}$ changed compared to that of FAB because of the different window partition methods.
The range of the relative position is $[-\mathbf{P} +\mathbf{P}^o-1,2\mathbf{P}^o-3]$; then, $\widehat{B}\in \mathbb{R}^{(P_x+P^{o}_x-1)\times (P_y+P^{o}_y-1)\times (P_z+P^{o}_z-1)}$ is parameterized, and the values in $B^{o}$ are taken from $\widehat{B}$.
OAB can enlarge the receptive field when calculating $key/value$ and promote the utilization of effective information from $query$, further reducing the optimization time cost and providing better visual representation.
Note that the local window cross-attention (LWCA) module proposed in~\cite{transmatch} used a cross-attention mechanism in medical image registration.
$query$ and $key/value$ computed in the LWCA module are from different images (fixed and moving images), respectively, while OAB calculates them inside each window with a larger field.
\subsection{Superresolution Upsampling}
The SR modules are employed to replace the interpolation upsampling layer and skip-connection operation to overcome image degradation in the encoder.
Several previous works~\cite{pixeljbhi1,pixeljbhi2,pixeltmi} in medical image analysis used the SR module to decrease subpixel location errors when the LR features are upsampled to HR features.
An SR module includes a channel convolutional layer, a pixel-shuffle layer~\cite{pixel}, a skip-connection operation and two convolutional layers.
The channel of the 3D feature map $N_{i+1} \in \mathbb{R}^{C \times H \times W \times D}$ is magnified by the channel convolutional layer by $l^{3}C$ times, and $l$ is the upsampling factor.
It is formulated as
\begin{equation}h=\text{Conv}(N_{i+1}),\label{eq12}\end{equation}
where Conv is the channel convolutional layer and $h$ is its output $h\in \mathbb{R}^{l^{3}C \times H \times W \times D}$.
Multiple features are rearranged by the pixel-shuffle layer into HR features $\widehat{h}\in \mathbb{R}^{C \times lH \times lW \times lD}$.
It is shown in Fig.\ref{fig3} and is written as
\begin{equation}\widehat{h}=\text{PixelShuffle}(h,l),\label{eq13}\end{equation}
The features in the encoder and HR features are concatenated via the skip-connection operation, and the information is fully integrated via two convolutional layers:
\begin{equation}N_i=\text{Conv}(\text{Conv}(\text{Cat}(\widehat{h},B_i))),\label{eq14}\end{equation}
where $B_i$ is the feature obtained by OAB in the encoder.
$N_i$ shows the fused features.
\subsection{Loss Functions}
The loss functions include three parts:
\begin{figure}[!t]
	\centerline{\includegraphics[width=\columnwidth]{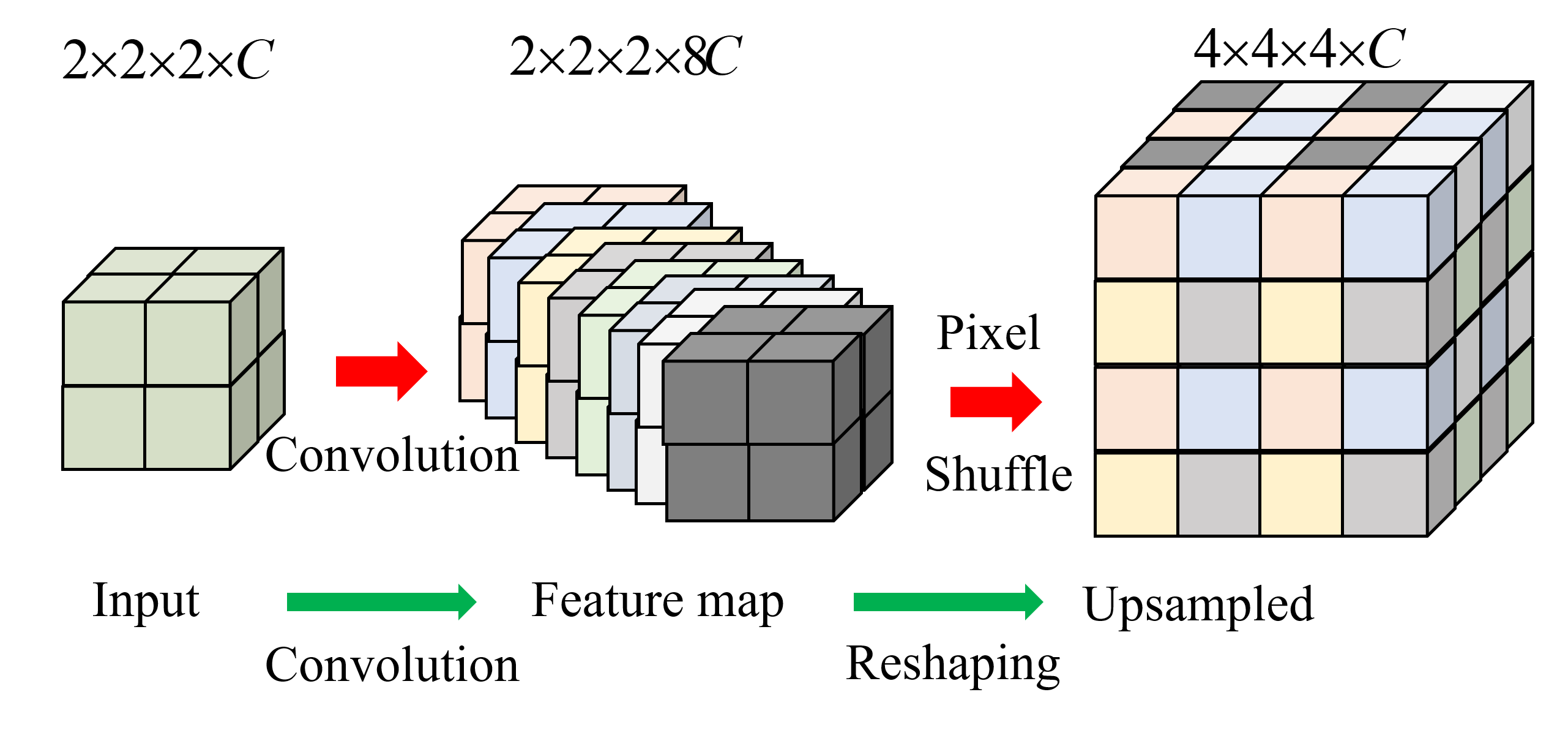}}
	\caption{Schematic diagram of the pixel-shuffle layer in SR module.}
	\label{fig3}
\end{figure}
The image similarity loss measures the similarity between a fixed image and a warped moving image.
The deformation field regularization loss promotes a smoother deformation field.
The improvement in the registration accuracy of auxiliary segmentation loss relies on the segmentation information of the major organs.
According to Eq. \ref{eq1}, the overall loss is formulated as
\begin{equation}\begin{split}\mathcal{L}(f,m,\phi)=\mathcal{L}_{sim}(f,m \circ \phi)+\lambda\mathcal{L}_{smooth}(\phi)\\
		+\gamma\{\mathcal{L}_{seg}(f_s,m_s\circ\phi)\}_{\text{optional}},\label{eq15}\end{split}\end{equation}
where $f_s$ and $m_s$ are the segmentation masks of the fixed and moving images, respectively.
$\lambda$ and $\gamma$ are the regularization parameter and auxiliary segmentation weight, respectively.
Note that $\mathcal{L}_{seg}$ is optional.
%It utilizes segmentation prior information and annotation knowledge of expert surgeons to achieve better performance~\cite{voxelmorphtmi}.
\subsubsection{Image Similarity Loss}
$\mathcal{L}_{sim}$ consists of a local normalized cross-correlation metric (LNCC) and mutual information (MI) voxelwise between $f$ and $m\circ \phi$.
The LNCC is calculated as
\begin{equation}\begin{split}&LNCC(f,m,\phi)=\\
&\sum_{\mathbf{p}\in \Omega}\frac{\left (\sum_{\mathbf{p}_i}\bigtriangleup f(\mathbf{p}_i)([m\circ \phi](\mathbf{p}_i)-[\hat{m}\circ \phi](\mathbf{p}))\right)^2}{\left(\sum_{\mathbf{p}_i}\bigtriangleup f^2(\mathbf{p}_i)\right)\left(\sum_{\mathbf{p}_i}([m\circ \phi](\mathbf{p}_i)-[\hat{m}\circ \phi](\mathbf{p}))^2\right)},\label{eq17}\end{split}\end{equation}
where $\mathbf{p}_i$ iterates over a local 3D cube centered at voxel $\mathbf{p}$ and the cube length is 9 in these experiments.
$\bigtriangleup f(\mathbf{p}_i) = f(\mathbf{p}_i) - \hat{f}(\mathbf{p})$.
$\hat{f}(\mathbf{p})=\frac{1}{9^3}\sum_{\mathbf{p}_i}f(\mathbf{p}_i)$ is the image with local mean intensities and $\hat{m}$ is same.
Since a higher LNCC represents a better registration accuracy, the loss function is $\mathcal{L}_{sim}=-LNCC$.

The mutual information (MI) between two random variables $X,Y$ can be defined as follows,
\begin{equation}MI(X,Y)=\iint p(x,y)\log\frac{p(x,y)}{p(x)p(y)}dxdy,\label{eq16}\end{equation}
where $p(x,y),p(x),p(y)$ are probability density functions. 
MI is known for its difficulty of accurate calculation in continuous high-dimensional spaces. 
To address this, researchers have proposed various approaches to estimate and optimize its calculation. 
For instance, the Parzen window density estimation method~\cite{KwakN} is utilized to derive the probability density and subsequently calculate MI. 
MI is frequently adopted for cross-modality-related tasks due to the relevance of their statistical features.
\subsubsection{Deformation Field Regularization Loss}
Optimizing $\mathcal{L}_{sim}$ may cause $\phi$ acute deformation with unrealistic physical motion.
Therefore, a diffusion regularizer is defined as
\begin{equation}\mathcal{L}_{smooth}(\phi)=\sum_{\mathbf{p}\in \Omega}||\bigtriangledown\mathbf{u}(\mathbf{p})||^2,\label{eq18}\end{equation}
where $\mathbf{u}$ is the spatial gradient of the displacement field.
\subsubsection{Auxiliary Segmentation Loss}
The segmentation information can encourage image registration during training but is not needed during testing.
The pixel labels used for image segmentation are annotated by professional surgeons or automatic segmentation algorithms.
The loss is formulated as
\begin{equation}\begin{split}&\mathcal{L}_{seg}(f_s,m_s\circ\phi)=\\&1-\frac{1}{K}\sum_{k=1}^{K}\frac{2\sum_{\mathbf{p}\in \Omega}f_s^k(\mathbf{p})[m_s^k\circ\phi](\mathbf{p})}{\sum_{\mathbf{p\in\Omega}}(f_s^k(\mathbf{p}))^2+\sum_{\mathbf{p\in\Omega}}([m_s^k\circ\phi](\mathbf{p}))^2},\label{eq19}\end{split}\end{equation}
where $f_s^k$ and $m_s^k$ are the binary mask of annotated structure/organ $k \in [1,K]$ for $f$ and $m$, respectively.
Notably, our proposed UTSRMorph is an unsupervised framework and auxiliary information can be optionally included to improve registration accuracy during training, such as segmentation labels.
\section{Experiments}
\subsection{Datasets}
We evaluated our proposed method in datasets including brain, abdomen and CMF tumor.
(1) OASIS~\cite{oasis} contains 413 T1W brain MR images and comes from the 2021 Learn2Reg challenge~\cite{Learn2Reg} for interpatient registration.
The original MR volumes were preprocessed by the organizing committee using FreeSurfer~\cite{freeSurfer} includes skull stripping, spatial normalization, affine transformations, and automatic structural segmentation.
A total of 413 T1W brain MR images were split into 394 and 19 images for the training set and validation set, respectively, which is similar to TransMorph~\cite{transmorph}.
Since the test set was not available, the validation set was used for direct evaluation.
The volumes were cropped to 160 $\times$ 192 $\times$ 224.
35 anatomical structures were used as ground truths to measure the registration accuracy.
(2) IXI~\cite{cycle} contains 576 T1W brain MR images.
The preprocessing operation was the same as that for the OASIS dataset.
%IXI imaging is an atlas-to-patient brain MR registration task.
The 576 T1 MR brain volumes were split into 403, 58 and 115 for the training, validation and testing sets, respectively.
The images were cropped to 160 $\times$ 192 $\times$ 224.
30 annotated structures were used for evaluation.
(3) Abdomen MR-CT~\cite{Learn2Reg}, a benchmark dataset for CT-MR registration in the Learn2Reg challenge, includes unpaired images of 40 MR images, 50 CT images for training and 8 MR-CT paired images for testing.
The organizers have labeled 3D segmentation masks for the liver, spleen, left and right kidney, which can provide auxiliary information to enhance registration accuracy during training. 
The target area for registration was cropped by the given regions of interest (ROI) and padded to 192 $\times$ 160 $\times$ 192, and then all images were aligned with a template image through affine transformation during preprocessing.
(4) Our CMF tumor MR-CT dataset contains 28 image pairs.
The age range was 23-88 years.
The shapes of the CT images were 514$\times$513$\times$512, and the spacings were 0.38$\times$0.43$\times$0.43 to 0.52$\times$0.6$\times$0.6 mm.
The MR volumes ranged from 24$\times$213$\times$256 to 244$\times$512$\times$512, and the spacings ranged from 0.9$\times$0.43$\times$0.43 to 6$\times$1.1$\times$1.1 mm.
The regions of interest patches were cropped according to the size of the MRI, then normalized, and affine transformation was used to convert them to the same coordinate system.
The input size was 128$\times$192$\times$224.
Since the facial tumors were small and the number of segmented organs was relatively small,
9 landmarks from our previous works~\cite{segmentation,HanBoxuan} were annotated by professional surgeons to evaluate registration performance, including the anterior nasal spine (ANS), upper incisor midpoint (UI), menton (Me), lowest point of the sigmoid notch on the right (SN-R), lowest point of the sigmoid notch on the left (SN-L), mandibular angle on the right (MA-R), mandibular angle on the left (MA-L), condyle on the right (Co-R), and condyle on the left (Co-L).
In each training iteration, we randomly selected two different volumes from the training set as the fixed and moving images in the brain or abdomen datasets. The CMF tumor dataset was trained by a pair of MR-CT images.
The annotated paired images were input into the trained network during the testing stage.
The study was approved by the ethics committee of Peking University School and Hospital of Stomatology (reference number: 2022-03).
\subsection{Evaluation Metrics}
In this subsection, several evaluation metrics~\cite{transmorph} are introduced.
(1) The Dice score (DSC) measures the overlap of masks with the same organ of warped moving and fixed images.
(2) The Hausdorff distance (HD95) describes the similarity of the anatomical segmentation between the fixed image and the registration result.
HD95 represents the 95th percentile.
(3) The Jacobian matrix indicates the deformation field quality.
We calculated the percentage of voxels with a nonpositive Jacobian determinant ($\%|J_\phi\leq0|$) or the standard deviation of the logarithm of the Jacobian determinant (SDlogJ).
(4) Target registration error (TRE) represents the Euclidean distance, which is calculated between the ground-truth and registered landmarks.
(5) Wilcoxon rank-sum test~\cite{DattaSomnath} is a standard procedure to test the equality of two distributions which is widely applied in statistics community. 
The advantage of Wilcoxon rank-sum test lies in its weak assumption on the underlying data distribution. 
The method is used to evaluate the significant differences between our proposed UTSRMorph and the baseline methods.
(6) Time cost (s) represents the inference time cost for testing a pair of images by the models.
(7) GPU Memory Usage (MB).
\subsection{Comparison Methods}
We compared our method with several other SOTA registration methods:
(1) SyN~\cite{syn}, the publicly available Python library ANTsPy was employed, and its transformation type was set to 'SyNOnly'.
The iteration numbers were set to (160, 80, 40).
(2) VoxelMorph~\cite{voxelmorphtmi}.
(3) ViT-V-Net~\cite{vitvnet}, the batch size is set to 1 in the abdomen MR-CT and OASIS datasets with segmentation loss, and the batch size of other datasets is set to 2. 
Note that in the GPU memory usage comparison experiment for ViT-V-Net, the batch size is set to 1 for fair comparison.
(4) LapIRN~\cite{MokTonyCW, MokTonyCWC}, the latest improved method with conditional instance normalization is used.
(5) ConvexAdam~\cite{SiebertHanna}, its feature extractor selects hand-crafted modality independent neighbourhood descriptor (MIND)~\cite{MIND} features and the total number of iterations is set to 100.
(6) NiftyReg~\cite{ModatMarc}.
LapIRN, ConvexAdam and NiftyReg are only evaluated in the OASIS and abdomen MR-CT datasets since they were specially designed for the two datasets and obtained excellent performance in 2021 Learn2Reg challenge.
(7) TransMorph~\cite{transmorph}.
(8) TransMatch~\cite{transmatch},
$\lambda=4$, and the learning rate was set to 0.0004.
(9) RDP~\cite{rdp}.
The number of RDP channels must be set to 12 in the OASIS, IXI and 10 in the MR-CT datasets due to GPU memory limitations.
All the methods are trained: 
in the IXI dataset using the LNCC loss and L2 smoothed loss with $\lambda=1$;
in the abdomen MR-CT and CMF tumor MR-CT datasets using MI loss and L2 smoothed loss with $\lambda=1$. 
Additionally, the segmentation loss is included in the abdomen MR-CT dataset with $\lambda=\gamma=1$; 
in the OASIS dataset using the LNCC loss and L2 smoothed loss with and without the segmentation loss, $\lambda=\gamma=1$.
The batch size is set to 1 and the learning rate is 0.0001.

\subsection{Implementation Details}
\begin{table}
	\caption{The results of all methods in the test set of the IXI dataset}
	\label{table}
	\setlength{\tabcolsep}{3pt}
	\begin{tabular}{p{50pt}p{47pt}p{40pt}p{40pt}p{15pt}p{15pt}}
		\hline
		Method & DSC(\%)& HD95& \%$|J_\theta|\leq 0$& Time& Memory\\
		\hline
		ANTs(SyN)& 64.45$\pm$15.19& 6.36$\pm$3.98& $\mathbf{0}$& 113.12&  \\
		ViT-V-Net& 73.49$\pm$12.59& 3.72$\pm$2.47& 1.61$\pm$0.35& $\mathbf{0.22}$& $\mathbf{11423}$\\
		VoxelMorph& 72.82$\pm$12.74& 3.89$\pm$2.69& 1.63$\pm$0.33& 0.23& 16028\\
		TransMorph& 75.11$\pm$12.35& 3.6$\pm$2.64& 1.47$\pm$0.32& $\mathbf{0.22}$& 17667\\
		RDP& 75.47$\pm$12.76& 3.66$\pm$2.82& 0.05$\pm$0.02& 0.9& 23911\\
		TransMatch& 76.24$\pm$12.53& $\mathbf{3.01}$$\pm$$\mathbf{2.12}$& 0.16$\pm$0.08& $\mathbf{0.22}$& 21985\\
		\hline
		UTSRMorph& $\mathbf{76.69}$$\pm$$\mathbf{12.56}$& 3.03$\pm$2.18& 0.32$\pm$0.13& 0.29& 19146\\
		UTSRMorph-L& 76.27$\pm$12.36& 3.03$\pm$2.09& 0.26$\pm$0.11& 0.52& 23187\\
		UTSRMorph-S& 76.07$\pm$12.7& 3.08$\pm$2.14& 0.27$\pm$0.11& $\mathbf{0.22}$& 18552\\
		\hline
		\multicolumn{6}{p{251pt}}{}
	\end{tabular}
	\label{tab1}
\end{table}
Three workstations with NVIDIA GeForceRTX 3090Ti or NVIDIA GeForceRTX 4090 GPUs were used to train different datasets.
The methods were built in Ubuntu 20.04, PyTorch 1.13.1 and Python 3.8.
The Adam optimizer was used to train the methods, and the batch size was 1.
The learning rates of OASIS and IXI were 0.0001 and 0.0004, respectively, and the maximum number of epochs was 500.
We used LNCC, $\mathcal{L}_{smooth}$ to train OASIS and IXI, and $\mathcal{L}_{seg}$ was optionally employed in OASIS.
$\lambda, \gamma$ was 1, 1 in OASIS, and $\lambda$ was 4 in IXI.
All methods were trained on the MR-CT datasets by MI loss.
$\mathcal{L}_{seg}$ was used to train in the abdomen dataset, $\lambda=\gamma=1$.
The learning rates of the abdomen and CMF tumor MR-CT datasets were 0.0002 and 0.0001, respectively, and the training was limited to 200,000 and 100,000 iterations, respectively.
The weighted factor $\alpha$ in FAB, channel compression parameter $\beta$ in the CA module and overlapping factor $\epsilon$ in OAB are 0.01, 3 and 0.5, respectively.
The window size of OAB was 4, and the head number was 4.
We proposed different architecture hyperparameters for the method variants, including UTSRMorph-S, UTSRMorph, and UTSRMorph-L.
The embedding dimensions were 48, 96, and 128; the numbers of FABs in the third stage were 2, 4, and 18; and the numbers of FABs in the other stages were 2.
The head numbers were \{4, 4, 4, 4\}, \{4, 4, 8, 8\} and \{4, 4, 8, 16\}.
\subsection{Results}
\subsubsection{Comparison with SOTA methods}
The SOTA methods with their proposed optimal training parameters were trained in the brain datasets (OASIS, IXI); the results are shown in Table~\ref{tab1},~\ref{tab2}.
\begin{table}
	\caption{The results of all methods in the test set of the OASIS dataset}
	\label{table}
	\setlength{\tabcolsep}{2pt}
	\begin{tabular}{p{50pt}p{50pt}p{40pt}p{40pt}p{15pt}p{15pt}}
		\hline
		Method & DSC(\%)& HD95& SDlogJ& Time& Memory\\
		\hline
		Initial& 57.18$\pm$21.26& 3.83$\pm$2.09& & & \\
		ANTs(SyN) & 70.53$\pm$36.09& 2.56$\pm$2.45& $\mathbf{0.05}$& 75& \\
		ViT-V-Net& 79.56$\pm$16.15& 2.12$\pm$2.08& 0.11$\pm$0.01& 1.17& 10703\\
		VoxelMorph& 78.65$\pm$16.8& 2.24$\pm$2.02& 0.13$\pm$0.01& 1.14& 10223\\
		NiftyReg& 78.84$\pm$16.31& 2.04$\pm$1.75&0.07& 107.78& \\
		LapIRN& 80.71$\pm$15.19& 1.94$\pm$1.9& 0.07& 1.2& 12477\\
		ConvexAdam& 74.27$\pm$17.61& 2.41$\pm$1.98& 0.13& 1.8& $\mathbf{3677}$\\
		TransMorph& 80.59$\pm$15.1& 1.97$\pm$1.84& 0.1$\pm$0.01&  1.14& 9811\\
		RDP& 81.17$\pm$14.45& $\mathbf{1.94}$$\pm$$\mathbf{1.76}$& 0.08& 1.74& 23275\\
		TransMatch& 79.01$\pm$16.08& 2.1$\pm$1.84& 0.06& 1.28& 21922\\
		\hline
		UTSRMorph& 81.06$\pm$14.37& 1.95$\pm$1.81& 0.1$\pm$0.01& 1.23& 15710\\
		UTSRMorph-L& $\mathbf{81.42}$$\pm$$\mathbf{14.19}$& $\mathbf{1.94}$$\pm$$\mathbf{1.89}$& 0.1$\pm$0.01& 1.41& 24139\\
		UTSRMorph-S& 80.36$\pm$15.08& 1.98$\pm$1.78& 0.11$\pm$0.01& $\mathbf{1.13}$& 9693\\
		\hline
		\multicolumn{6}{p{251pt}}{To reduce GPU usage of TransMatch and maintain consistency during training, the shallowest convolutional block of TransMatch, TransMorph and UTSRMorph were abandoned.}
	\end{tabular}
	\label{tab2}
\end{table}
Our proposed UTSRMorph achieved the optimal DSC and HD95 while having a lower SDlogJ, $\%|J_\phi\leq0|$.
This finding indicates that UTSRMorph can generate high-quality, detailed deformation fields and achieve excellent registration performance.
Moreover, the method demonstrated competitive results in terms of GPU memory usage and inference time.
Among the other comparison methods, the registration performances of ViT-V-Net, TransMorph and TransMatch are superior to that of VoxelMorph due to the long-distance dependency of the Transformer, especially the feature representation learning ability of Swin-T.
The cross-attention calculation between a pair of input images occupies considerable GPU memory in TransMatch, which causes the convolution blocks to be abandoned.
This limits its performance.
RDP, which uses a recursive pyramid strategy to focus on the deformation fusion block and upsampling layer in the decoder, achieved the closest registration result to ours; however, the deformation fusion block specifically occupied GPU memory.
According to the above methods, our method fully considers the feature extraction ability of the encoder for leveraging the advantages of ConvNets and Swin-T, and uses the SR module in the upsampling layer of the decoder to balance the smoothness and detail deformation field generating.
\begin{table}
	\caption{The results of all methods in the OASIS dataset with segmentation loss}
	\label{table}
	\setlength{\tabcolsep}{2pt}
	%\centering
	\begin{tabular}{p{40pt}p{50pt}p{40pt}p{40pt}p{20pt}p{20pt}}
		\hline
		Method & DSC(\%)& HD95& SDlogJ& Time& Memory\\
		\hline
		ViT-V-Net& 84.48$\pm$11.78& 1.66$\pm$1.93& $\mathbf{0.12}$$\pm$$\mathbf{0.01}$& $\mathbf{1.11}$& 22455\\
		VoxelMorph& 84.18$\pm$12.19& 1.67$\pm$1.75& 0.12$\pm$0.02& 1.18& $\mathbf{20616}$\\
		TransMorph& 85.86$\pm$10.63& 1.46$\pm$1.43& 0.12$\pm$0.02&  1.14& 21207\\
		\hline
		UTSRMorph& $\mathbf{86.28}$$\pm$$\mathbf{10.22}$& $\mathbf{1.42}$$\pm$$\mathbf{1.4}$& 0.13$\pm$0.02& 1.23& 23337\\
		\hline
		\multicolumn{6}{p{251pt}}{RDP and TransMatch with segmentation loss were excluded from comparative methods due to high GPU memory usage.}
	\end{tabular}
	\label{tab3}
\end{table}
Then, segmentation loss was added to train these models in OASIS; the results are shown in Table~\ref{tab3}.
The auxiliary segmentation information can promote global matching information correspondence by facilitating local information matching of pivotal organs during training.
Although it can significantly improve registration accuracy, the segmentation loss calculation for several different categories also prominently increases computational resources, resulting in some methods with large GPU memory being out-of-commission.
\begin{figure}[!t]
	\centerline{\includegraphics[width=\columnwidth]{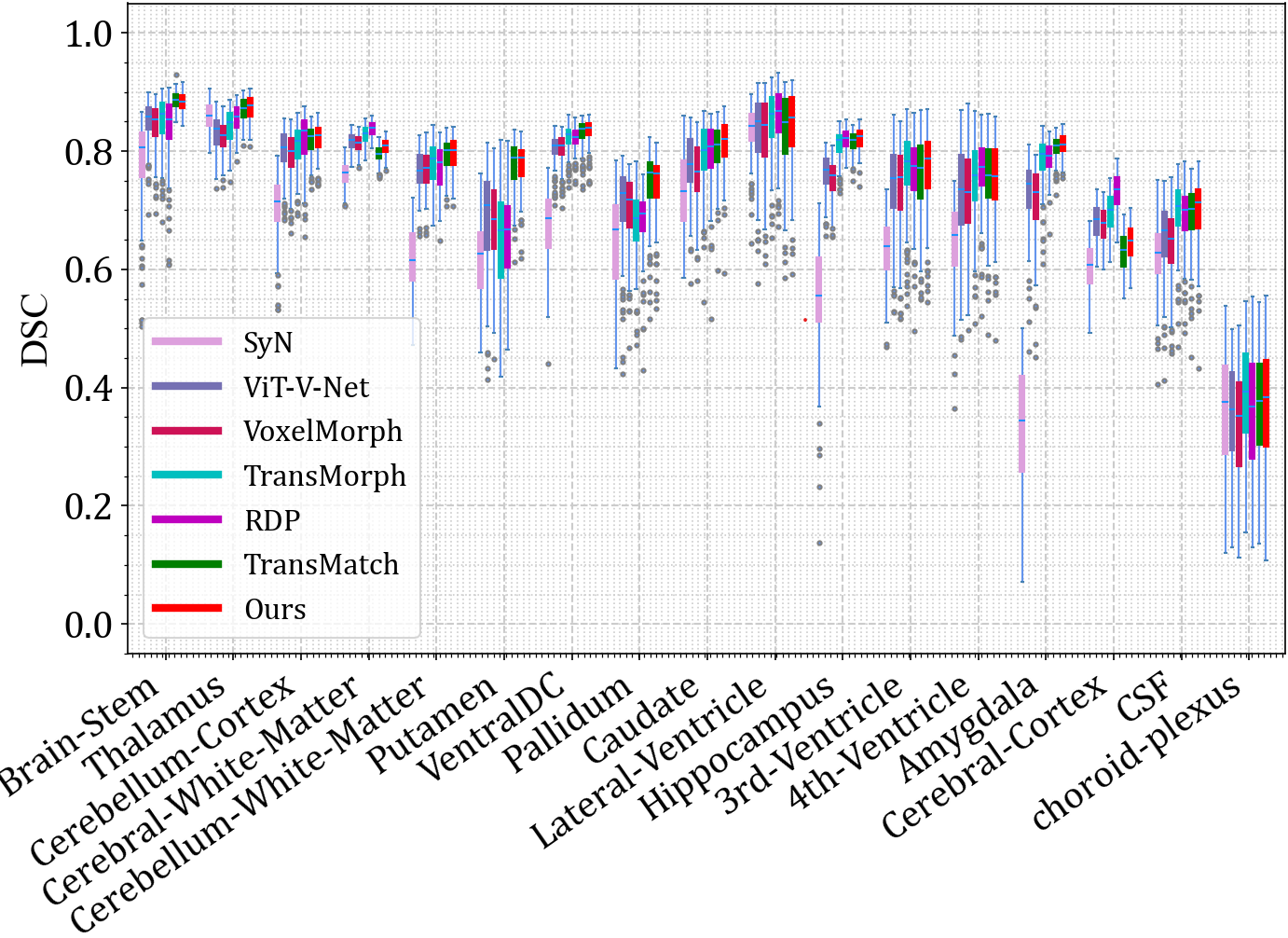}}
\caption{Boxplots with Dice scores of various registration methods in the IXI dataset.}
	\label{fig4}
\end{figure}
The detailed DSC distributions obtained with different methods for various organs are shown in Fig.\ref{fig4}. The optimal DSC of most organs was acquired for UTSRMorph.
\begin{table}
	\caption{The results of all methods in the test set of the abdomen MR-CT dataset}
	\label{table}
	\setlength{\tabcolsep}{2pt}
	\begin{tabular}{p{50pt}p{50pt}p{38pt}p{40pt}p{15pt}p{15pt}}
		\hline
		Method & DSC(\%)& HD95& SDlogJ& Time& Memory\\
		\hline
		Initial& 40.27$\pm$16.86 &   & & & \\
		Affine & 57.24$\pm$19.42& & & & \\
		ViT-V-Net& 69.07$\pm$18.58& 9.29$\pm$5.99& 0.03$\pm$0.01& 0.23& 14838\\
		VoxelMorph& 67.81$\pm$18.5& 10.23$\pm$6.24& 0.02$\pm$0.01& 0.11& 13975\\
		NiftyReg& 72.61$\pm$22.89& 8.54$\pm$8.34&$\mathbf{0}$& 117.8& \\
		LapIRN& 70.91$\pm$19.3& 10.48$\pm$6.05& 0.1$\pm$0.01& 0.22& 13457\\
		ConvexAdam& $\mathbf{77.16}$$\pm$$\mathbf{19.12}$& $\mathbf{7.63}$$\pm$$\mathbf{7.63}$& 0.07& 0.39& $\mathbf{3086}$\\
		TransMorph& 70.97$\pm$20.68& 10.64$\pm$7.96& 0.04$\pm$0.01&0.52& 19608\\
		RDP& 76.77$\pm$22.54& 7.81$\pm$6.02& 0.05& 0.26& 23893\\
		TransMatch& 70.45$\pm$20.74& 11.12$\pm$8.96& 0.02$\pm$0.01& 0.14& 23492\\
		\hline
		UTSRMorph& 75.22$\pm$18.7& 9.72$\pm$7.77& 0.03$\pm$0.01& 0.11& 17229\\
		UTSRMorph-L& 73.7$\pm$19.31& 9.81$\pm$7.62& 0.03$\pm$0.01& 0.17& 23011\\
		UTSRMorph-S& 73.18$\pm$19.24& 9.57$\pm$7.86& 0.02& $\mathbf{0.09}$& 13475\\
		\hline
		\multicolumn{6}{p{251pt}}{The convolutional blocks of TransMatch and UTSRMorph were abandoned.}
	\end{tabular}
	\label{tab4}
\end{table}

\begin{table}
	\caption{The unsupervised results of all methods in the test set of the abdomen MR-CT dataset}
	\label{table}
	\setlength{\tabcolsep}{2pt}
	\begin{tabular}{p{50pt}p{50pt}p{38pt}p{40pt}p{15pt}p{15pt}}
		\hline
		Method & DSC(\%)& HD95& SDlogJ& Time& Memory\\
		\hline
		ViT-V-Net& 65.59$\pm$19.74& 10.15$\pm$7.11& 0.03$\pm$0.01& 0.23& 14036\\
		VoxelMorph& 65.71$\pm$19.73& 10.42$\pm$6.99& 0.02$\pm$0.01& $\mathbf{0.11}$& $\mathbf{13063}$\\
		TransMorph& 68.76$\pm$20.54& 9.92$\pm$7.24& 0.03$\pm$0.01&0.52& 18282\\
		RDP& 64.81$\pm$19.99& 10.43$\pm$7.09& $\mathbf{0}$& 0.26& 22276\\
		TransMatch& 68.75$\pm$19.53& 9.99$\pm$7.16& 0.02$\pm$0.01& 0.14& 21433\\
		\hline
		UTSRMorph& $\mathbf{69.89}$$\pm$$\mathbf{19.39}$& $\mathbf{9.58}$$\pm$$\mathbf{7.06}$& 0.02$\pm$0.01& $\mathbf{0.11}$& 16004\\
		\hline
		\multicolumn{6}{p{251pt}}{The convolutional blocks of TransMatch and UTSRMorph were abandoned.}
	\end{tabular}
	\label{tab7}
\end{table}

\begin{figure}[!t]
	\centerline{\includegraphics[width=\columnwidth]{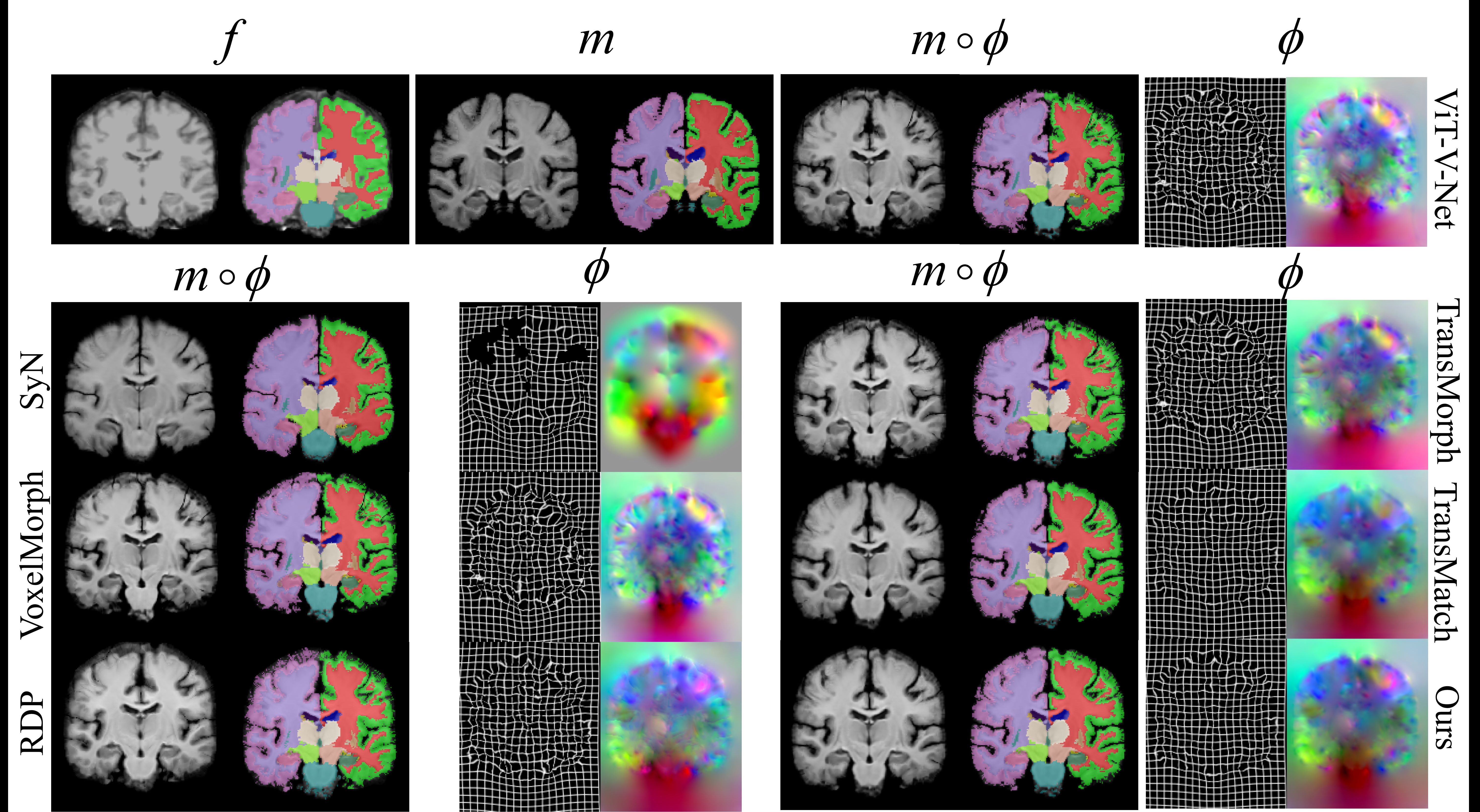}}
\caption{The visualization results of various registration methods in the IXI dataset.
		The results include the raw image and ground-truth mask of the moving and fixed images, warped image and warped mask, estimated continuous and RGB deformations.The RGB deformation is generated by mapping each spatial dimension of the displacement field to a channel of the RGB color.}
	\label{fig5}
\end{figure}

The visualization results for the IXI dataset are shown in Fig.\ref{fig5}.
A deformation field with local excessive distortion and missing detail was obtained by the traditional method, resulting in poor accuracy.
The deformation field obtained by our method is generally smooth, behaves well, and maintains almost no folding.
The segmentation results after image registration indicate that our method can consistently preserve internal structures by balancing local deformation and global smoothness using the advantages of ConvNets and ViTs, especially for organs with smaller volumes.

\begin{figure}[!t]
	\centerline{\includegraphics[width=\columnwidth]{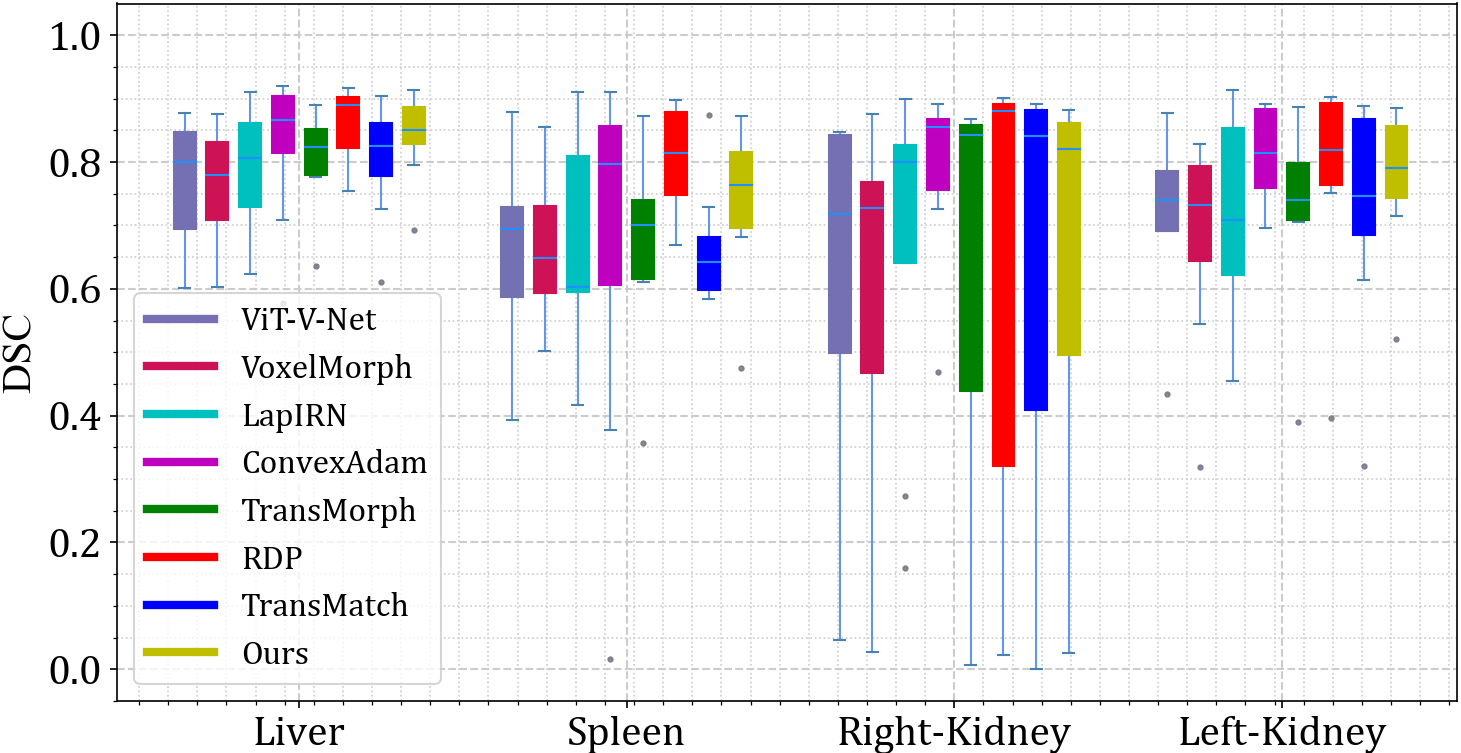}}
\caption{Boxplots with Dice scores of various registration methods in the abdomen MR-CT dataset.}
	\label{fig6}
\end{figure}
To achieve better performance, the segmentation loss was employed to train the abdomen MR-CT dataset.
The SOTA methods were compared in the abdomen MR-CT dataset; the results are shown in Table~\ref{tab4} and Fig.\ref{fig6}.
The optimal performance is achieved by ConvexAdam.
\begin{table}
	\caption{The results of all methods in the test set of the craniomaxillofacial tumors MR-CT dataset}
	\label{table}
	\setlength{\tabcolsep}{3pt}
	\begin{tabular}{p{50pt}p{50pt}p{40pt}p{45pt}p{35pt}}
		\hline
		Method & TRE (mm)& SDlogJ& Time& Memory\\
		\hline
		Initial& 15.85$\pm$9.29 &   & &  \\
		ViT-V-Net& 6.03$\pm$3.22& 0.02$\pm$0.01& 0.77& 13277\\
		VoxelMorph& 5.9$\pm$3.43& 0.01$\pm$0.01& 0.77& 12965\\
		TransMorph& 5.63$\pm$3.15& 0.01$\pm$0.01& $\mathbf{0.75}$&13656\\
		RDP& 5.61$\pm$3.25& $\mathbf{0}$& 0.93& 22198\\
		TransMatch& 5.77$\pm$3.1& 0.01$\pm$0.01& 1.32& 20519\\
		\hline
		UTSRMorph& $\mathbf{5.54}$$\pm$$\mathbf{3.33}$& 0.01& 0.81& 15888\\
		UTSRMorph-L& 5.71$\pm$3.05& 0.01$\pm$0.01& 0.9& 21144\\
		UTSRMorph-S& 5.72$\pm$3.53& 0.01& 0.77&  $\mathbf{12279}$\\
		\hline
		\multicolumn{5}{p{251pt}}{The shallowest convolutional block of TransMatch, TransMorph and UTSRMorph were abandoned.}
	\end{tabular}
	\label{tab5}
\end{table}
Our method achieved an accuracy which was second only to that of RDP and significantly greater than that of the other deep learn-based methods.
The unsupervised results are presented in Table~\ref{tab7}. 
After discarding the auxiliary segmentation information, the performance of all unsupervised methods drops noticeably. 
Our method achieves better Dice scores, and surprisingly, the performance of RDP decreases significantly.
The evaluation metrics of our datasets are shown in Table~\ref{tab5}; our method achieved an optimal TRE index and generated relatively smooth deformation fields.
\begin{figure}[!t]
	\centerline{\includegraphics[width=\columnwidth]{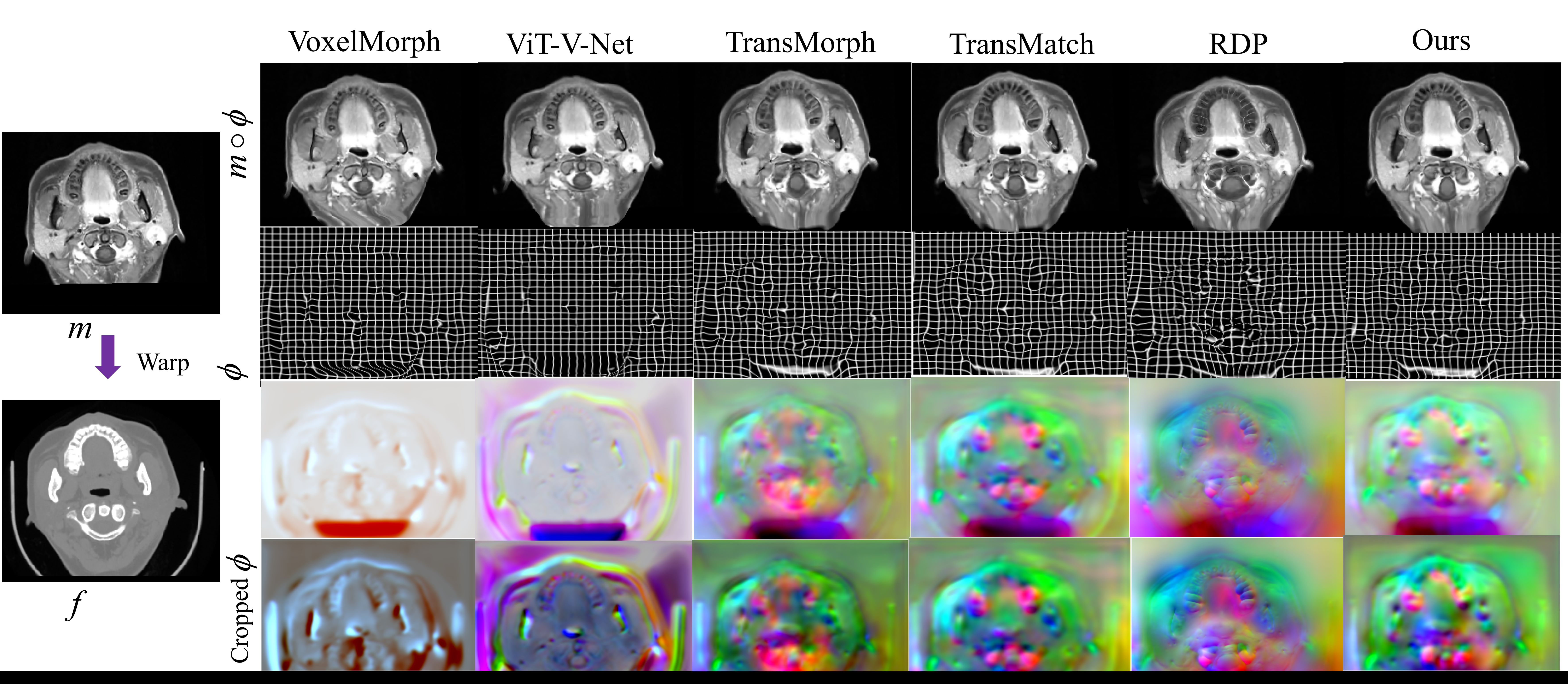}}
\caption{The visualization results of various registration methods in our CMF tumor MR-CT dataset.
		The moving image is MR volume and the fixed image is CT volume.
		The $\phi$ is cropped to show that MR volume is a subset of CT volume.}
	\label{fig8}
\end{figure}
The visualization results of the more detailed displacement field and warped moving image are shown in Fig.\ref{fig8}.
The local deformation in the deformation field of the RDP is more pronounced when adapting to larger shape changes; the local spatial location in the displacement field is highlighted.
This causes local regional distortion.
Our method utilizes local and global information to generate a smoother deformation field; the local detail feature in a warped moving image conforms to the fixed image.
\subsubsection{Ablation Experiment}
\begin{table}
	\caption{The ablation experiment of all methods for proposed modules}
	\label{table}
	\setlength{\tabcolsep}{2pt}
	\begin{tabular}{p{30pt}|p{20pt}p{10pt}p{15pt}p{15pt}|p{50pt}p{40pt}p{40pt}}
		\hline
		Dataset& $\mathcal{L}_{seg}$& SR& FAB& OAB& DSC& HD95& SDlogJ\\
		\hline
		\multirow{4}{*}{OASIS}& \multirow{4}{*}{\ding{52}}& \ding{56}& \ding{56}& \ding{56}& 85.86$\pm$10.63& 1.46$\pm$1.43& 0.12$\pm$0.02\\
		& & \ding{52}& \ding{56}& \ding{56}& 85.98$\pm$10.53&1.46$\pm$1.44&$\mathbf{0.12}$$\pm$$\mathbf{0.01}$\\
		& & \ding{52}& \ding{52}& \ding{56}& 86.07$\pm$10.45&1.47$\pm$1.46&0.12$\pm$0.02\\
		& & \ding{52}& \ding{52}& \ding{52}& $\mathbf{86.28}$$\pm$$\mathbf{10.22}$&$\mathbf{1.42}$$\pm$$\mathbf{1.4}$&0.13$\pm$0.02\\
		\hline
		Dataset& $\mathcal{L}_{seg}$& SR& FAB& OAB& DSC& HD95& \%$|J_\theta|\leq 0$\\
		\hline
		\multirow{4}{*}{IXI}& \multirow{4}{*}{\ding{56}}& \ding{56}& \ding{56}& \ding{56}& 75.11$\pm$12.35& 3.6$\pm$2.64& 1.47$\pm$0.32\\
		& & \ding{52}& \ding{56}& \ding{56}& 76.37$\pm$12.51&3.07$\pm$2.15&$\mathbf{0.3}$$\pm$$\mathbf{0.12}$\\
		& & \ding{52}& \ding{52}& \ding{56}& 76.38$\pm$12.57&3.05$\pm$2.2&0.32$\pm$0.14\\
		& & \ding{52}& \ding{52}& \ding{52}& $\mathbf{76.69}$$\pm$$\mathbf{12.56}$& $\mathbf{3.03}$$\pm$$\mathbf{2.18}$& 0.32$\pm$0.13\\
		\hline
		Dataset& $\mathcal{L}_{seg}$& SR& FAB& OAB& DSC& HD95& SDlogJ\\
		\hline
		\multirow{4}{*}{\makecell[c]{Abdomen\\MR-CT}}& \multirow{4}{*}{\ding{52}}& \ding{56}& \ding{56}& \ding{56}& 70.97$\pm$20.68& 10.64$\pm$7.96& 0.04$\pm$0.01\\
		& & \ding{52}& \ding{56}& \ding{56}& 72.6$\pm$20.31&10.29$\pm$7.65&$\mathbf{0.02}$$\pm$$\mathbf{0.01}$\\
		& & \ding{52}& \ding{52}& \ding{56}& 74.29$\pm$18.56&$\mathbf{9.1}$$\pm$$\mathbf{7.47}$&0.03$\pm$0.01\\
		& & \ding{52}& \ding{52}& \ding{52}& $\mathbf{75.22}$$\pm$$\mathbf{18.7}$&9.72$\pm$7.77&0.03$\pm$0.01\\
		\hline
		\multicolumn{8}{p{251pt}}{\ding{52} represents this module used in UTSRMorph and \ding{56} indicates abandoned module.}
	\end{tabular}
	\label{tab6}
\end{table}
To evaluate the effectiveness of each module in improving accuracy, we conducted ablation experiments where we omitted the OAB, replaced the FAB with Swin-T, and replaced the SR modules with interpolation upsampling layers.
The results of the ablation experiments of UTSRMorph in different datasets are shown in Table~\ref{tab6}.
Our proposed SR, FAB and OAB can gradually increase the DSC and decrease the distortion of the deformation field.
This finding indicates that these modules can enhance the feature representation ability and generate more accurate deformation fields, further improving the corresponding transformation between all points of a pair of images.
The SR module decreases $\%|J_\theta|\leq 0$, which demonstrates that the module can improve the smoothness of a displacement field and generate a more detailed displacement field.
Although FAB and OAB can significantly increase DSC, the rate of unrealistic deformation also slightly increases.
This proves its feature extraction and representation learning ability.
\begin{figure}[!t]
	\centerline{\includegraphics[width=\columnwidth]{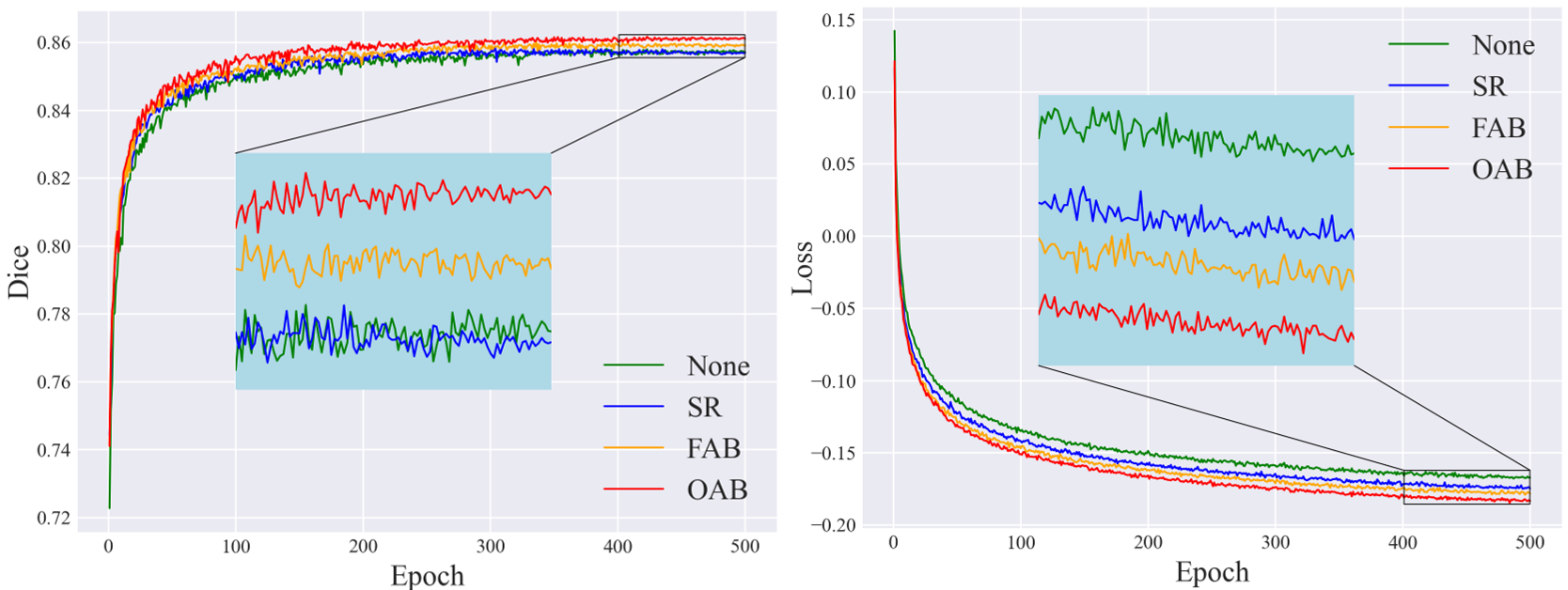}}
\caption{The trend of change with Dice scores and loss in the test set of OASIS trained by segmentation loss.
		The curves named none, SR, FAB, OAB represent adding each module gradually, as shown in Table~\ref{tab6}.}
	\label{fig7}
\end{figure}
A typical change trend with DSCs and loss in the test set of OASIS trained by segmentation loss is shown in Fig.\ref{fig7}.
The curve between the number of iterations and the loss decreases faster when our proposed SR, FAB and OAB are added.
This indicates that aggregating ConvNets and ViTs accelerates the error optimization speed significantly.
ConvNets can reduce the redundancy of ViTs and accelerate optimization during training.
The optimal DSCs obviously increase when OAB using the cross-attention module is added.
It indicates the cross-window can integrate abundant correlations with visual information better.
\section{Discussion}
The distributions of DSCs in the IXI dataset are shown in Fig.\ref{fig4}. 
Our UTSRMorph achieves the best registration performance across most organs, including thalamus, cerebellum-white-matter, ventralDC, caudate, hippocampus, 3rd-ventricle, amygdala, CSF.
In other organs, our method gains competitive results comparable to TransMatch or RDP. 
Specifically, RDP obtains higher DSCs in organs with significant deformations, such as cerebellum-cortex and cerebral-cortex, highlighting its effectiveness in handling large deformations. 
TransMatch enhances registration performance in organs with complex structures, such as the brain-stem and putamen, by leveraging multiple features. 
Large-sized organs (such as cortex, ventricle, abdomen organs) frequently require significant deformations, while small-sized organs (such as thalamus, hippocampus, putamen) have complex relationships with surrounding tissues. 
However, our UTSRMorph can aggregate rich features extracted by ConvNets and ViTs and enlarge the receptive field using the overlapping attention modules, which can provide a better balance between handling large deformations and addressing intricate local structures. 
The DSCs and registration performance of choroid-plexus are worse in all methods. 
The reason may be due to complex inter-attachment with surrounding organs and diffuse morphological structures.

The distributions of DSCs in the abdomen MR-CT dataset are illustrated in Fig.\ref{fig6}. 
The accuracy (especially for right-kidney) of the networks based entirely on deep learning is lower than the ConvexAdam, which utilizes hand-crafted MIND features and thin plate splines. 
The reason may be that the learning-based approaches using segmentation loss was subject to significant overfitting on the liver, spleen and left-kidney. 
The overfitting results led to unsatisfactory results on the right-kidney. 
Our UTSRMorph demonstrates competitive results comparable to RDP which excels in handling large deformations. 
Except that, our method significantly outperforms other learning-based networks. 
This finding indicates our network's capability to effectively compensate for the large deformation.

The statistical analysis of paired Wilcoxon rank-sum test in the all datasets between UTSRMorph and other baseline methods for the DSCs is conducted. 
The $p$ value of UTSRMorph compared with other baseline methods in the IXI dataset is  $p<0.05$. 
The $p$ value in the OASIS dataset is $p<0.5$ except for RDP ($p=0.28$) and TransMorph ($p=0.22$). 
The reason may be that the testing set of the OASIS dataset is small (19 samples) while the total number of the testing set in the IXI dataset is 115 samples. 
For the results in the OASIS dataset with segmentation loss, the performance of UTSRMorph is not significantly different from TransMorph ($p=0.11$). 
The reason may be that the methods using segmentation loss was subject to significant overfitting. 
Although the abdomen MR-CT and CMF tumor MR-CT datasets are not suitable for statistical analysis because of their small sample sizes (8 samples and 4 samples, respectively), we still made the statistical tests. 
The $p$ values of LapIRN, ConvexAdam and RDP are 0.07, 0.41 and 0.09 in the abdomen MR-CT dataset trained with segmentation loss and the performance of UTSRMorph is significantly different from other methods. 
The results in the abdomen MR-CT dataset show no significant difference between all unsupervised methods when discarding the auxiliary segmentation information. 
The Dice scores of UTSRMorph are not significantly different in the CMF tumor MR-CT datasets. 
Besides the impact of test data size, the errors of the annotating landmarks, particularly in MR images with large spacing (over 2mm), also play a significant role. 
In subsequent work we will collect more MR data with smaller spacing and provide denser landmark annotations to improve the effectiveness of TRE index. 
For most entries, the $p$ value is less than 0.05, indicating that the DSCs improvement of our proposed method is significant.
\begin{table}
	\caption{The comparison results of different upsampling layers in IXI dataset}
	\label{table}
	\setlength{\tabcolsep}{2pt}
	\begin{tabular}{p{80pt}p{55pt}p{55pt}p{55pt}}
		\hline
		Upsampling type & DSC(\%)& HD95& \%$|J_\theta|\leq 0$\\
		\hline
		Cubic interpolation& 76.43$\pm$12.57& 3.07$\pm$2.18& 0.4$\pm$0.1\\
		TransConv& 76.56$\pm$12.26& 3.05$\pm$2.18& 0.4$\pm$0.1\\
		SR& $\mathbf{76.69}$$\pm$$\mathbf{12.56}$& $\mathbf{3.03}$$\pm$$\mathbf{2.18}$& $\mathbf{0.3}$$\pm$$\mathbf{0.1}$\\
		\hline
		\multicolumn{4}{p{251pt}}{}
	\end{tabular}
	\label{tab8}
\end{table}

\begin{figure}[!t]
	\centerline{\includegraphics[width=\columnwidth]{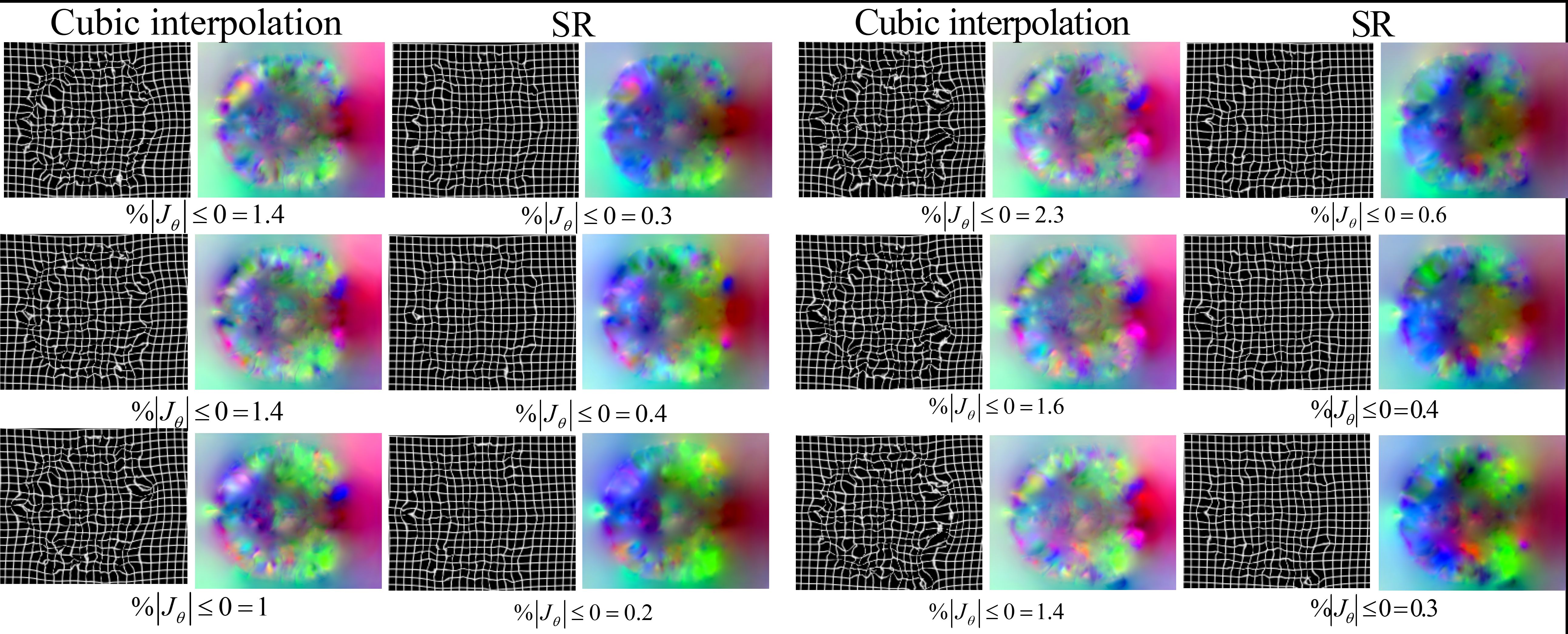}}
	\caption{The visualization of the deformation field and the corresponding Jacobian determinant for the same slices between using our SR modules and using the cubic interpolation upsampling layers for TransMorph in the IXI dataset.}
	\label{fig9}
\end{figure}

\begin{figure}[!t]
	\centerline{\includegraphics[width=\columnwidth]{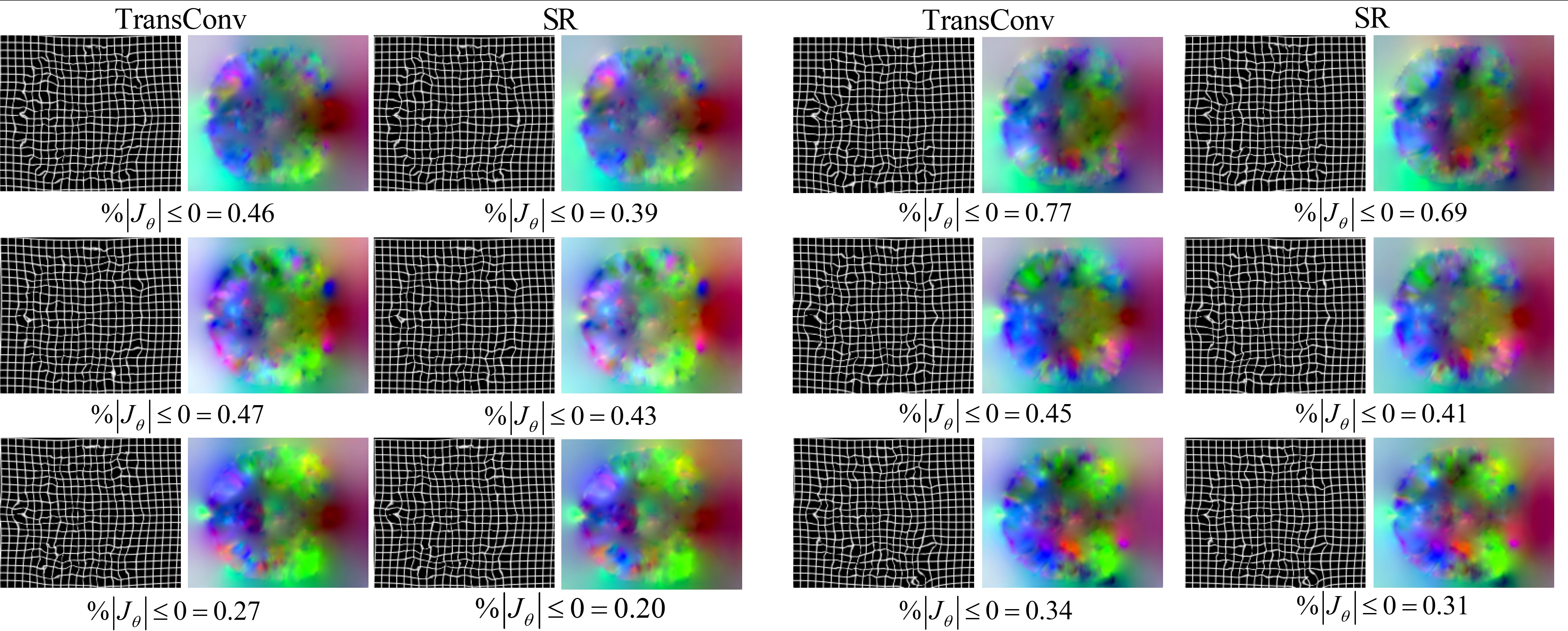}}
	\caption{ The visualization of the deformation field and the corresponding Jacobian determinant for the same slices between using our SR modules and using the TransConv layers for our UTSRMorph in the IXI dataset.}
	\label{fig10}
\end{figure}

To demonstrate the effectiveness of the SR module, we have conducted additional experiments as follows. 
We replaced the original interpolated upsampling layer with the SR module and investigated the performance improvement in TransMorph.
The compared results of the displacement field in the IXI dataset are visualized in Fig.\ref{fig9}. 
These visualizations clearly demonstrate that displacement fields generated by the SR module exhibit smoother deformations with no local folding of patches. 
Moreover, a lower nonpositive Jacobian determinant further confirms the advantages of the SR module. 
This finding indicates the SR module can enhance displacement smoothness and gradually upscale the features to produce more detailed displacement fields.
We compared cubic interpolation, TransConv, and SR at the same time in our UTSRMorph and the results are shown in Table~\ref{tab8}. 
The SR module slightly outperforms the others.
The visualization results of TransConv and SR are shown in Fig.\ref{fig10}.
To measure the differences of deformation field quality between our method and the cubic interpolation upsampling layers in Fig.\ref{fig9}, TransConv layers in Fig.\ref{fig10}, the statistical test results of the Jacobian matrix index are $p<10^{-3}$, $p=0.01$.
The results indicate that the deformation field produced by the SR module achieves smoother results compared to the other two upsampling layers.

In this paper, we have observed that SR, a common research field for the ill-posed image inverse reconstruction, plays a crucial role in enhancing registration performance. 
It has not been investigated in the image registration field and is a first attempt in this work.
However, the pixel shuffle layer is subject to repeating artifacts as going deeper, whose parameters are difficult to learn in deeper layers~\cite{moser}.
This may reduce the accuracy of deeper UTSRMorph-L.
In the future work, we will continue to focus on the SR to improve the accuracy of image registration. 
We hope that integrating more SR concepts into the deformable image registration community will not only enhance the detail in displacement fields but also mitigate feature degradation.

\section{Conclusion}
In summary, we proposed UTSRMorph to generate a deformation field for medical image registration tasks.
It fully utilizes the advantages of ConvNets and ViTs to retain local features while reducing the long-distance information redundancy of global dependency.
Our method uses a cross-attention mechanism that relies on overlapping windows to enhance the representation learning ability of the encoder.
In addition, the SR module can improve the local deformation accuracy while ensuring the smoothness of the generated deformation field.
Our method achieved competitive results in four datasets. Additionally, the experimental results validated its effectiveness in multimodal registration tasks.

\appendices

\bibliographystyle{IEEEtran}
\begin{center}
\bibliography{reference}
\end{center}

\end{document}